\documentclass[10pt,twocolumn,letterpaper]{article}

\usepackage{iccv}
\usepackage{times}
\usepackage{epsfig}
\usepackage{graphicx}
\usepackage{amsmath}
\usepackage{amssymb}
\usepackage{bbm}

\usepackage{algorithm}
\usepackage{algorithmic}
\usepackage{booktabs}

\DeclareMathOperator*{\argmax}{arg\,max}

\usepackage[pagebackref=true,breaklinks=true,letterpaper=true,colorlinks,bookmarks=false]{hyperref}

\iccvfinalcopy 

\def\iccvPaperID{1307} 

\ificcvfinal\fi
\begin{document}

\title{Point Linking Network for Object Detection}

\author{Xinggang Wang$^1$, Kaibing Chen$^1$, Zilong Huang$^1$, Cong Yao$^2$, Wenyu Liu$^1$\\
$^1$School of EIC, Huazhong University of Science and Technology, $^2$Megvii Technology Inc.\\
{\tt\small \{xgwang, kbchen, hzl, liuwy\}@hust.edu.cn, yaocong@megvii.com}
}

\maketitle

\begin{abstract}
Object detection is a core problem in computer vision. With the development of deep ConvNets, the performance of object detectors has been dramatically improved. The deep ConvNets based object detectors mainly focus on regressing the coordinates of bounding box, \eg, Faster-R-CNN, YOLO and SSD. Different from these methods that considering bounding box as a whole, we propose a novel object bounding box representation using points and links and implemented using deep ConvNets, termed as Point Linking Network (PLN). Specifically, we regress the corner/center points of bounding-box and their links using a fully convolutional network; then we map the corner points and their links back to multiple bounding boxes; finally an object detection result is obtained by fusing the multiple bounding boxes. PLN is naturally robust to object occlusion and flexible to object scale variation and aspect ratio variation. In the experiments, PLN with the Inception-v2 model achieves state-of-the-art single-model and single-scale results on the PASCAL VOC 2007, the PASCAL VOC 2012 and the COCO detection benchmarks without bells and whistles. The source code will be released.
\end{abstract}

\section{Introduction}

Object detection is one of the most fundamental problems in computer vision. It has various real-world applications, ranging from robotics, autonomous car, to video surveillance and image retrieval. Object detection is very challenging since it suffers from scale variation, viewpoint change, intra-class variation, shape variation, and occlusion of object, as well as background clutters. Deep ConvNets are excellent at learning representation, thus, to some extent, deep ConvNets based object detectors are more robust to the challenges mentioned above. However, explicit modeling is still beneficial to solving the object detection problem. Different from the previous deep ConvNets based detectors which focus on learning better features for this task, in this paper, we propose to study better object model to achieve better object detector in the framework of deep ConvNets.

\begin{figure}
  \centering
  \includegraphics[width=0.95\linewidth]{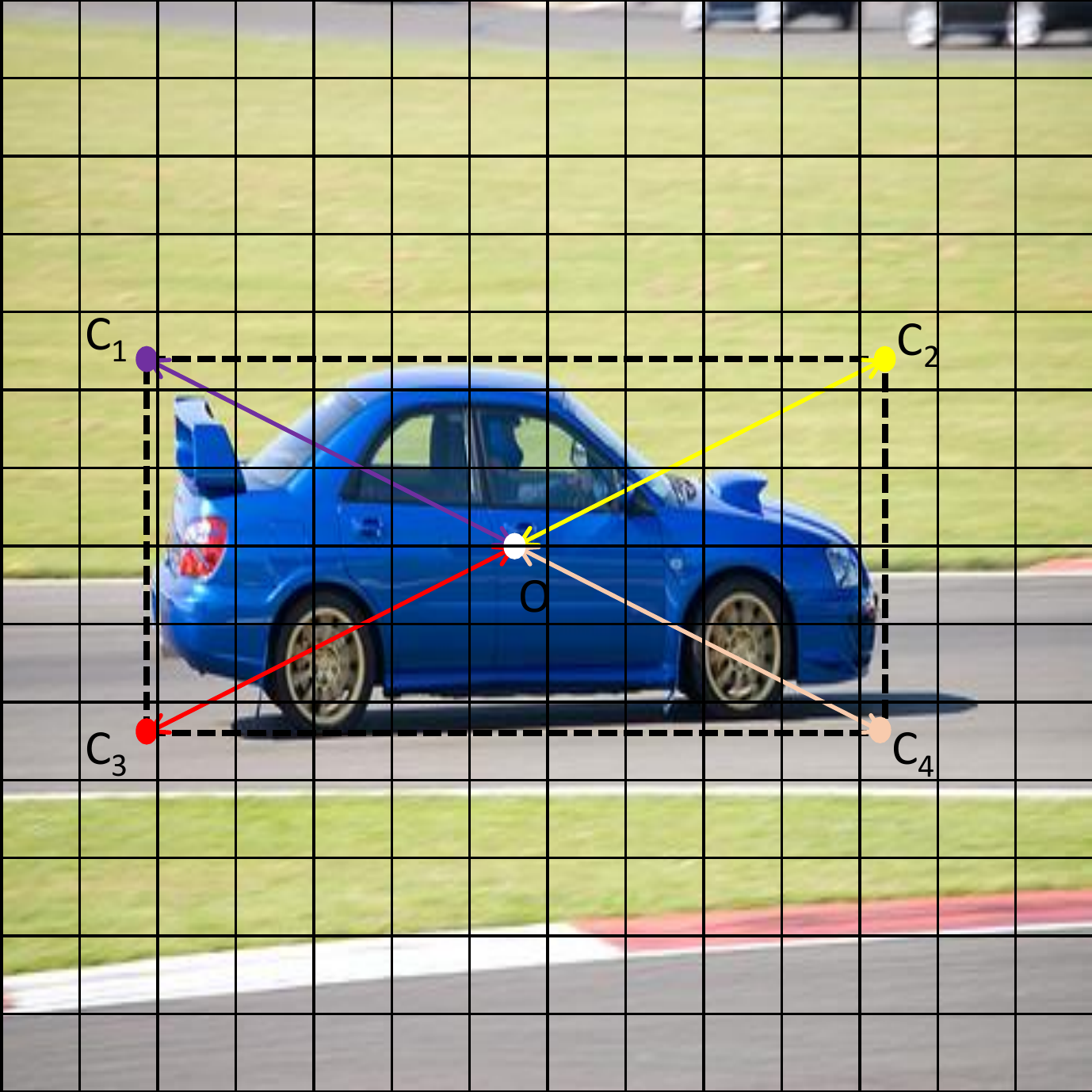}\\
  \caption{Illustration of our object detection idea using point linking. Our algorithm detects object by predicting the center-corner point pair, such as $OC_1$, $OC_2$, $OC_3$, and $OC_4$. The positions of points will be predicted according to the grids in the image. Once we get a pair of center and corner points, the bounding box of object is easily obtained.}    \label{fig:idea}
\end{figure}

After the deep ConvNets had obtained great successes on image classification~\cite{Ref:Krizhevsky2012}, utilizing deep ConvNets for object detection is becoming a new challenge. The developments of object detection using deep learning can be summarized into two stages. In the first stage, the object detectors based on deep ConvNets follows the ``sliding window" strategy. They learn classifiers to check candidate windows one by one and obtain the top scoring candidate windows after non-maximum suppression as the detection results, such as OverFeat \cite{sermanet2013overfeat}, R-CNN \cite{girshick2014rich}, Fast-R-CNN \cite{girshick2014rich}. These kinds of methods have some limitations:  1) The detector only sees local information when making a decision and is hard to reason about global context. 2) Even most of the deep CNN features of candidate windows are shareable, learning additional features for individual windows and classifying them are still computationally expensive. Thus, these methods are relatively slow. In the second stage, deep detectors can directly predict/regress object bounding boxes using a single ConvNet, such as Deep MultiBox~\cite{erhan2014scalable}, YOLO~\cite{Ref:YOLO-Redmon2016}, Faster-R-CNN~\cite{Ref:FasterRCNN-Ren2015}, and~SSD \cite{Ref:SSD-Redmon2016}. A common strategy is to map a labeled bounding-box to a grid in the convolutional feature map and then each cell responses to predicting the coordinates and class label (\eg, $x, y, w, h, c$) of the mapped bounding box. When predicting the coordinates, usually a cell has multiple anchors for different default aspect ratios of bounding box. The crux of these methods is representing a various number of bounding boxes using a fixed length vector (the grids with no matched bounding box will be ignored during training). Deep ConvNets are excellent at learning a mapping between the input image and a fixed length vector contains its semantic information and there is no extra computation for a large number of candidate windows. Thus YOLO, Faster-R-CNN, SSD \etc have obtained very impressive object detection accuracy and speed.

However, current bounding box regression methods still suffer from the large variation of the aspect ratios of bounding boxes, scale variation, and occlusion. To address these problems, we propose a new solution for object detection based deep networks. We cast the object detection problem into two sub-problems: a point detection problem and a point linking problem. As shown in Fig.~\ref{fig:idea}, there are two kinds of points in our system, the center point of an object bounding box which is denoted as $O$ and a corner point of an object bounding box, \eg $C_A$. There are four corner points in Fig.~\ref{fig:idea} which are the top-left corner, the top-right corner, the bottom-left corner and the bottom-right corner. It is clear that any pair of center point and corner point can determine a bounding box. In order to detect the point pair, there are two tasks, the first task is to localize the two points that is termed as point detection, and the second task is to associate the two points, say the points belonging to the same object, which termed as point linking. This is the general object detection idea in this paper, which is point-based object detection framework. Compared to the previous bounding box regression methods, this new framework has three advantages: 1) It is very flexible to represent bounding box in any scale and any aspect ratio. 2) For each bounding box, there are at least four pairs of points, object detection performance can be boosted via voting. 3) It is naturally robust to occlusion, since it can infer object location using local cues.

We implement this point-based object detection framework using a single deep network, termed as Point Linking Network (PLN). In PLN, both point detection and point linking are implemented in a joint loss function. As shown in Fig.~\ref{fig:idea}, each grid cell represents a grid cell the convolutional feature map of the input image. A grid cell is responsible for predicting the center point inside includes its confidence, x-offset, y-offset and link,  as well as the corner point inside also includes its confidence, x-offset, y-offset and link. The details will be introduced in Section~\ref{sec:method}. We have tested it on the standard object detection benchmarks, PASCAL VOC 2007 \& 2012,and COCO. The results show PLN is superior to the widely used Faster-RNN, YOLO and SSD in the same data augmentation setting.


\section{Related Work}

Recently, object detection has become one of the most active and productive areas in computer vision, mainly driven by two factors: deep convolutional neural networks (CNN)~\cite{Ref:LeCun1990, Ref:Krizhevsky2012} and large-scale image datasets~\cite{Ref:ImageNet-Deng2009, Ref:MSCOCO-Lin2014}. Numerous insightful ideas and effective algorithms have been proposed, since the emergence of the pioneering work R-CNN~\cite{Ref:RCNN-Girshick2014}, which developed an object detection paradigm that first finds object proposals and successively classifies each proposal with a trained CNN model.

To improve R-CNN, He~\etal~\cite{Ref:SPPNet-He2014} presented SPP-net, which adopts a spatial pyramid pooling layer to generate a fixed-length representation regardless of image scale and aspect ratio. Girshick~\cite{Ref:FastRCNN-Girshick2015} proposed to process whole image with several convolutional and max pooling layers to produce feature maps and compute a fixed-length feature for each proposal via a RoI pooling layer. Later, Ren~\etal~\cite{Ref:FasterRCNN-Ren2015} introduced a Region Proposal Network (RPN) to replace the time-consuming region proposal computation procedure. The resultant model, called Faster R-CNN, achieves excellent performance on standard benchmarks for object detection while running much faster. Different from R-CNNs, the  proposed  PLN method is proposal free.

Different from the R-CNN paradigm, the YOLO framework~\cite{Ref:YOLO-Redmon2016} casts object detection as a regression problem to spatially separated bounding boxes and associated class probabilities. YOLO is able to run in real-time but makes more localization errors. YOLO9000~\cite{Ref:YOLO9000-Redmon2017}, as an extension to YOLO, made several modifications (such as batch normalization, high resolution classifier and direct location prediction) and outperformed prior arts for object detection while still running significantly faster. Alternatively, SSD~\cite{Ref:SSD-Redmon2016} completely eliminates proposal generation and subsequent feature re-sampling stages, by discretizing the output space of bounding boxes into a set of default boxes over different aspect ratios and scales and integrating predictions from multiple feature maps of different scales. However, PLN is able to generating bounding boxes in any scale and any aspect ration.

Making use of the inherent multi-scale, pyramidal hierarchy of deep convolutional networks, Lin~\cite{Ref:FeaturePyramid-Lin2016} constructed feature pyramids with a top-down architecture with lateral connections. This method obtains state-of-the-art accuracy on the COCO detection benchmark~\cite{Ref:MSCOCO-Lin2014}. Dai~\etal~\cite{Ref:R-FCN-Li2016} devised a region-based object detector that is fully convolutional with almost all computation shared. Position-sensitive score maps are used to handle different parts of objects. The ResNet~\cite{Ref:ResNet-He2016} has advanced image classification with identity mappings and much deeper networks, and it has proven to be very effective when superseding previous base model (for example, VGG-16~\cite{Ref:Simonyan2015}) in detection systems. These tricks also work with the proposed PLN method. In PLN, we choose the Inception-v2 network as  base network without any special design in the feature extraction layer.

PLN is related to some semantic segmentation methods, such as FCN~\cite{shelhamer2016fully} and Deeplab~\cite{chen2016deeplab}, since we  are trying to segment out the corner points and  center points. In addition, PLN is similar to problem of multi-person pose estimation problem \cite{insafutdinov16ariv, cao2016realtime}. However, PLN aims at object detection and is much faster than these pose estimation methods.

Before the era of deep learning, Deformable Part Model (DPM)~\cite{felzenszwalb2010object} has been the state-of-the-art object detector for a long period of time. Different from current deep object detectors, DPM is part based and robust to object deformation. In this paper, we draw inspiration from DPM to perform part-based object detection, within in the current deep learning based object detection framework.

\section{Point Linking Network for Object Detection}
\label{sec:method}

\subsection{Network Design for Object Detection}
\label{sec:design}

Following the setting of the single-shot object detectors \cite{Ref:SSD-Redmon2016, Ref:YOLO9000-Redmon2017}, we resize the input image $I$ into a fixed size with equal height and width.\footnote{PLN also works with images with different width and height. But for the simplicity of implementation and describing the method, we set width equal to height.}  The base network of PLN is Inception-v2 which is a fast and accurate network for object detection according to the survey paper proposed by Google \cite{huang2016speed}. The convolutional feature map of $I$, denoted as $F$, generated using Inception-v2 has the spatial dimension of $S\times S$. Then, we divide $I$ into $S\times S$ grid cells as shown in Fig.~\ref{fig:idea}. Thus, a grid cell in $I$ is associated with a grid cell in $F$. For $i \in [1, \cdots, S^2]$, $I_i$ denotes the $i$-th grid cell in image and $F_i$ denotes the $i$-th grid cell in $F$.

In PLN, each $F_i$ is responsible for $2\times B$ points predictions, which are consisted of $B$ center predictions and $B$ corner predictions. If we want to predict multiple object center/corner in a grid cell, we set $B>1$ . Without the loss of generalizability, we suppose the corner is a top-right corner. The $1$st to the $B$-th predictions are for center points and the $(B+1)$-th to the $(2B)$-th predictions are for corner points. Each prediction contains four items: $P_{ij}$, $Q_{ij}$, $\left[x_{ij}, y_{ij} \right]$, $\left[ L^x_{ij}, L^y_{ij} \right]$, where $i \in [1, \cdots, S^2]$ is the \textit{spatial index} and $j \in [1, \cdots, 2\times B]$ is the \textit{point index}. The meanings of the four items are explained as follows.

\begin{itemize}
\item $P_{ij}$ is the probability that a point exists in the grid cell; it has two forms, $P(O)_{ij}$ and $P(C)_{ij}$ denotes the existence of center point and corner point, respectively. 
\item $Q_{ij}$ is the probability distribution over object classes; suppose there are $N$ classes; $Q(n)_{ij}, n\in[1,\cdots,N],$ is the probability that the point belongs to the $n$-th class. 
\item $\left[x_{ij}, y_{ij}\right]$ represents the accurate position of the point which is relative to the top-right corner of the cell grid in $I$ and normalized by the size of grid cell. Thus, both $x_{ij}$ and $y_{ij}$ are both in the range of $[0, 1]$. 
\item $\left[L^x_{ij}, L^y_{ij}\right]$ denotes the link of the point. There is a special design. Both $L^x_{ij}$ and $L^y_{ij}$ are vectors in the length of $S$. The $k$-th element $L(k)^x_{ij}$ and $L(k)^y_{ij}$, $k\in[1,S]$, is the probability of the point linked to the grid cell in the $k$-th row and the $k$-th column, respectively. The probabilities are normalized to make sure $\Sigma_{k=0}^S L(k)^x_{ij} = 1$ and $\Sigma_{k=0}^S L(k)^y_{ij} = 1$. Thus, $[L^x_{ij}, L^y_{ij}]$ means the point indexed with $(i,j)$ is linked to the grid cell with the row index of $\argmax_k L(k)^x_{ij}$ and the column index of $\argmax_k L^y_{ij}$. Besides of the spatial index, if the point is a center point, $j \in [1,B]$, it links to the corner point with the point index of $(j+B)$; if the point is a corner point, $j \in [B+1,2B]$, it links to the center point with the point index of $(j-B)$. For simplicity, we use $\pi(L^x_{ij}, L^y_{ij})$ to denote the linked index which include both the spatial index and the point index.
\end{itemize}

\begin{figure*}[ht!]
  \centering
  \includegraphics[width=0.9\linewidth]{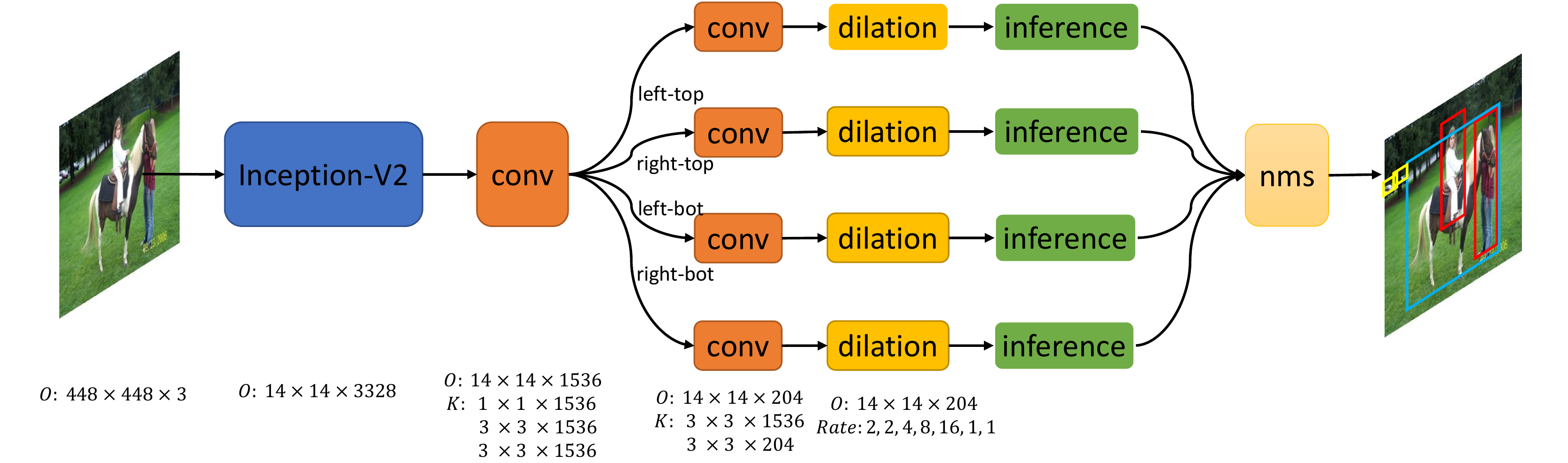}\\
  \caption{The network architecture of PLN. The detection network is based on Inception-v2. We use Inception-v2 and some additional convolutional layers to regress parameters of points, then parse the parameters to obtain the bounding box and category label of object. Finally, we combine boxes of four branch (left-top, right-top, left-bot and right-bot) and apply NMS to obtain final object detection result.}    
  \label{fig:network}
\end{figure*}

\subsection{Loss Function}
\label{sec:loss}

Based on the above definitions, we give the training loss of a point indexed with $(i,j)$. There are two cases, a grid cell contains a type of point or not, which are indicated using $\mathbbm{1}^\text{pt}_{ij}$ and $\mathbbm{1}^\text{nopt}_{ij}$. It does not matter the point inside is a center point or a corner point. If the cell grid indexed with $i$ and $j$ contains the point,  $\mathbbm{1}^\text{pt}_{ij} = 1$ and $\mathbbm{1}^\text{nopt}_{ij} = 0$; otherwise,  $\mathbbm{1}^\text{pt}_{ij}=0$ and $\mathbbm{1}^\text{nopt}_{ij}=1$. Thus, the loss function of this cell grid is given as follows:
\begin{equation}
\text{Loss}_{ij}  = \mathbbm{1}^\text{pt}_{ij} \text{Loss}^\text{pt}_{ij} + \mathbbm{1}^\text{nopt}_{ij} \text{Loss}^\text{nopt}_{ij}
\end{equation}

Then, if the grid cell contains a point, the grid cell needs to predict the existence, the class distribution probabilities, the precise location, and the linked point. The loss function $\text{Loss}^\text{pt}_{ij} $ is formulated as follows:
\begin{equation}
\begin{aligned}
\text{Loss}^\text{pt}_{ij} = (P_{ij} - 1)^2 + w_\text{class} \Sigma_{n=1}^N \left(
Q(n)_{ij} - \hat{Q(n)_{ij}}
\right)^2 + \\
w_\text{coord}\left( (x_{ij} - \hat{x}_{ij})^2 + (y_{ij} - \hat{y}_{ij})^2 \right) + \\
w_\text{link} \Sigma_{k=1}^S \left( (L(k)^x_{ij} - \hat{L(k)^x_{ij}})^2 + (L(k)^y_{ij} - \hat{L(k)^y_{ij}})^2 \right),
\end{aligned}
\end{equation}
in which, the variables with a hat are the ground truths, which are easily to be obtained from the original bounding box annotations in object detection benchmarks. Basically, we minimize the least squares errors of the point existence, classification scores, x-offset, y-offset, and rough location of the linked point. $w_\text{class}$, $w_\text{coord}$, and $w_\text{link}$ are the weight parameters for the point existence, the precise location, and the rough location of the linked point respectively.

Meanwhile, the loss function of a grid with no point inside is given as follows:
\begin{equation}
\text{Loss}^\text{nopt}_{ij} = P_{ij} ^ 2,
\end{equation}
which means the existence $P_{ij}$ should be approached to zero.

In summary, the final loss function of PLN is given as 
\begin{equation}
\text{Loss} = \sum_{i=1}^{S^2}\sum_{j=1}^{2B} L_{ij}.
\end{equation}
It is the sum over all grid cells and all types of points.
Notice that, in the proposed Loss, we simply use Euclidean loss to regress the predictions. From our experiences, we found out that the Euclidean loss in PLN is robust and it is not necessary to design different loss functions for different types of predictions.

\subsection{Inference}
\label{sec:inference}

Inferencing the object detection results from PLN is straight forward. For a pair of points indexed with $(i,j)$ and $(s,t)$, the probability of the point pair being an object of the $n$-th class is given by
\begin{equation}
\label{eq:inference}
\begin{aligned}
P^\text{obj}_{ijnst} = P_{ij} P_{st} Q(n)_{ij} Q(n)_{st}
\frac{ L(s_x)^x_{ij} L(s_y)^y_{ij}  + L(i_x)^x_{st} L(i_y)^y_{st} }{2}.
\end{aligned}
\end{equation}
In which, we decompose the spatial index ($i, s$) into its x-component and y-component respectively. $i_x$, $i_y$, $s_x$, and $s_y$ are all in the range of $[1, S]$. Recall that we have a hard constraint that a link can only be existed between a pair of center point and corner point, which is denoted as $|j -t|=B$). Thus, $L(s_x)^x_{ij} L(s_y)^y_{ij}$ means the probability point $(i,j)$ links to $(s,t)$  and $L(i_x)^x_{st} L(i_y)^y_{st}$ means the probability point $(s,t)$ links to $(i,j)$.

Eq.~\eqref{eq:inference} involves a problem with five loops and seems to be computationally expensive. However, 1) the numbers of the loops are relative small. In the experiments, we have  $S\leq 20$, $B=2$ and $N=20$. 2) There are some heuristics can be applied to reduce the computations. For example, we just need to search a top-right corner point in the top-right areas. Thus, the inference process can be quickly computed. 

\subsection{Branches Merging via NMS}

As shown in Fig.~\ref{fig:idea}, we can generate four center-corner pairs from an object bounding box. In PLN, we have four branches for regressing the object bounding box as shown in Fig.~\ref{fig:network}. The four branches sharing the Inception-v2 base network are individually trained using the loss function given in Section~\ref{sec:loss} and using the same ground truth. In testing, the four branches are inferred (Section~\ref{sec:inference}) individually. Finally, the inferred bounding boxes of the four branches are fed into the non-maximum suppression (NMS) process to obtain the final detection results. As explained in our introduction, the four branches merging method helps to improve object detection performance by handling the occlusion problem.

\section{Experiments}

We perform experiments on a large variety of object detection datasets including the PASCAL VOC 2007 dataset~\cite{everingham2007pascal}, the PASCAL VOC 2012 dataset, the People-Art dataset~\cite{cai2015cross}, and the COCO dataset~\cite{lin2014microsoft}.

\subsection{Implementation Details} 

We implement the proposed PLN detector in TensorFlow~\cite{abadi2016tensorflow}. The PLNs are mainly based on Inception-v2 \cite{ioffe2015batch} pre-trained on the ImageNet image classification dataset provided in TensorFlow\footnote{\url{http://download.tensorflow.org/models/inception_v2_2016_08_28.tar.gz}}. The Inception-v2 model of Tensorflow was trained using RMSProp optimizer, so we train PLNs using this optimizer as well. Specifically, we remove the final pooling layer, softmax layer and auxiliary branch of inception, then add some convolutional layers after the last inception block. Previous works have proven that adding some convolutional layers after ConvNet can improve detection performance. We thus use convolution layers to generate H $\times$ W feature map, and then use sigmoid and softmax operator to restrict the feature which represent probability of point existence, probability distribution over class, precise locations, and the link information to the matching point. Four branches are used to predict the point pairs of object, including (center, left-top), (center, right-top), (center, left-bottom) and (center, right-bottom). In the testing phase, we put all bounding boxes of four branches together, then apply non-maximum suppression to form the final object boxes.

We fine-tune the networks with an initial learning rate of 0.001, then slowly raise the learning rate from 0.001 to 0.005. We continue training with 0.005 for 30000 iterations.Throughout training, we use a batch size of 64, a momentum of 0.9, and a decay of 0.00004. We use a mount of data augmentation to train network. In training time, we randomly choose a simplified SSD data augmentation and a modified YOLO data augmentation. The simplified SSD data augmentation means that we only crop image patches applied in SSD but does not apply the photo-metric distortions. The modified YOLO data augmentation means that we change the jitter parameters from $[-0.2, 0.2]$ to $[-0.3, 0.1]$. The reason is because the SSD data augmentation usually crops image patch within the image, it tends to enlarge the objects. To remedy this issue, we change the jitter parameters in YOLO data augmentation to crop larger image patch outside of the original image to remedy it. 

We mainly focus on comparing PLN detector with three typical as well as the state-of-the-art deep detectors: Faster R-CNN~\cite{Ref:FasterRCNN-Ren2015}, YOLO~\cite{Ref:YOLO-Redmon2016}, and SSD\cite{Ref:SSD-Redmon2016}. YOLO and SSD are single-shot detectors; Faster R-CNN and YOLO are single-scale detector. Here, we refer SSD as a multi-scale detector since it uses multi-scale deep feature hierarchies to predict object bounding box. The PLN detector is single-shot and single-scale.

The source code of PLN will be released.

\subsection{PASCAL VOC 2007}

We comprehensively evaluate PLN on the PASCAL VOC 2007 dataset. The training data includes VOC 2007 trainval and  VOC 2012 trainval, which has 16551 images in total. The testing data is VOC 2007, which has 4952 images. On this dataset, we compare PLN with Fast R-CNN, Faster R-CNN, YOLO and SSD using the mean Average Precision (mAP) measure. 

Table~\ref{table:voc07} shows the results of Faster R-CNN, SSD, YOLOv2 and the PLN detectors with different sizes of input image. The sizes of input images are $448\times 448$, $512\times 512$ and $640\times 640$. The sizes of corresponding convolutional feature maps are $14$, $16$, and $20$ respectively.  By increasing the image sizes from 448 to 640, PLN has a very consistent performance improvement.
The mAP of PLN448 is similar to SSD512, is higher than SSD300 and Faster R-CNN. PLN512 is better than SSD512 which obviously shows the advantage of the proposed pointing linking detection loss function over the previous coordinate regression method. PLN640 obtains a mAP of 78.8\% which outperforms the previous state-of-the-art YOLOv2 by 0.2\% mAP. The sizes of input image of YOLOv2 is $544 \times 544 $, slightly smaller than PLN640. However, YOLOv2 utilizes a pre-trained model using high resolution ImageNet images, which is computationally expensive. Our PLN detectors do not use the high resolution pre-trained model and will also be beneficial from it.

\begin{table*}
\caption{{PASCAL VOC 2007 test detection results.} 
The image size for YOLOv2 is 544 $\times$ 544. Faster R-CNN restricts its input images has a minimum dimension of 600. The two SSD models have different input sizes of 300 $\times$ 300 and 512 $\times$ 512. The proposed PLN has the input sizes of 448 $\times$ 448, 512 $\times$ 512 and 640 $\times$ 640. }
  \centering
  \resizebox{\textwidth}{!}{\begin{tabular}{l|c|c c c c c c c c c c c c c c c c c c c c}
  \toprule
    Method & mAP & aero & bike & bird & boat & bottle & bus & car & cat & chair & cow & table & dog & horse & mbike & person & plant & sheep & sofa & train & tv \\
  \midrule
    Faster \cite{Ref:FasterRCNN-Ren2015}\footnote{Faster R-CNN uses ResNet-xx as the base model.}   & 76.4 & 79.8 & 80.7 & 76.2 & 68.3 & 55.9 & 85.1 & 85.3 & 89.8 & 56.7 & 87.8 & 69.4 & 88.3 & 88.9 & 80.9 & 78.4 & 41.7 & 78.6 & 79.8 & 85.3 & 72.0 \\
    SSD300 \cite{Ref:SSD-Redmon2016}    & 74.3 & 75.5 & 80.2 & 72.3 & 66.3 & 47.6 & 83.0 & 84.2 & 86.1 & 54.7 & 78.3 & 73.9 & 84.5 & 85.3 & 82.6 & 76.2 & 48.6 & 73.9 & 76.0 & 83.4 & 74.0 \\
    SSD512 \cite{Ref:SSD-Redmon2016}    & 76.8 & 82.4 & 84.7 & 78.4 & 73.8 & 53.2 & 86.2 & 87.5 & 86.0 & 57.8 & 83.1 & 70.2 & 84.9 & 85.2 & 83.9 & 79.7 & 50.3 & 77.9 & 73.9 & 82.5 & 75.3 \\
    YOLOv2 \cite{Ref:YOLO9000-Redmon2017}    & 78.6 &  -   &   -  &  -   &  -   &  -   &  -   &  -   &  -   &  -   &  -   &  -   &  -   &  -   &  -   &  -   &  -   &  -   &  -   &  -   &  -   \\
    \hline
    PLN448     & 76.8 & 80.8 & 85.9 & 76.7 & 64.8 & 56.4 & 84.6 & 82.8 & 88.7 & 60.8 & 80.6 & 71.7 & 86.7 & 85.6 & 83.2 & 76.8 & 52.4 & 80.9 & 76.5 & 86.0 & 74.4 \\
    PLN512     & 77.8 & 78.9 & 86.5 & 77.5 & 72.0 & 58.0 & 84.9 & 85.3 & 86.2 & 62.1 & 83.3 & 73.9 & 84.7 & 85.6 & 86.0 & 78.6 & 52.3 & 80.5 & 77.6 & 85.8 & 75.9 \\
    PLN640     & 78.8 & 83.4 & 85.8 & 80.4 & 68.6 & 63.7 & 86.3 & 86.2 & 88.0 & 61.4 & 85.5 & 72.7 & 86.3 & 87.8 & 85.6 & 81.0 & 50.8 & 84.4 & 75.9 & 85.8 & 76.4 \\
  \bottomrule
  \end{tabular}}
  \label{table:voc07}
\end{table*}

\begin{figure*}
  \centering
    \includegraphics[width=0.8\linewidth]{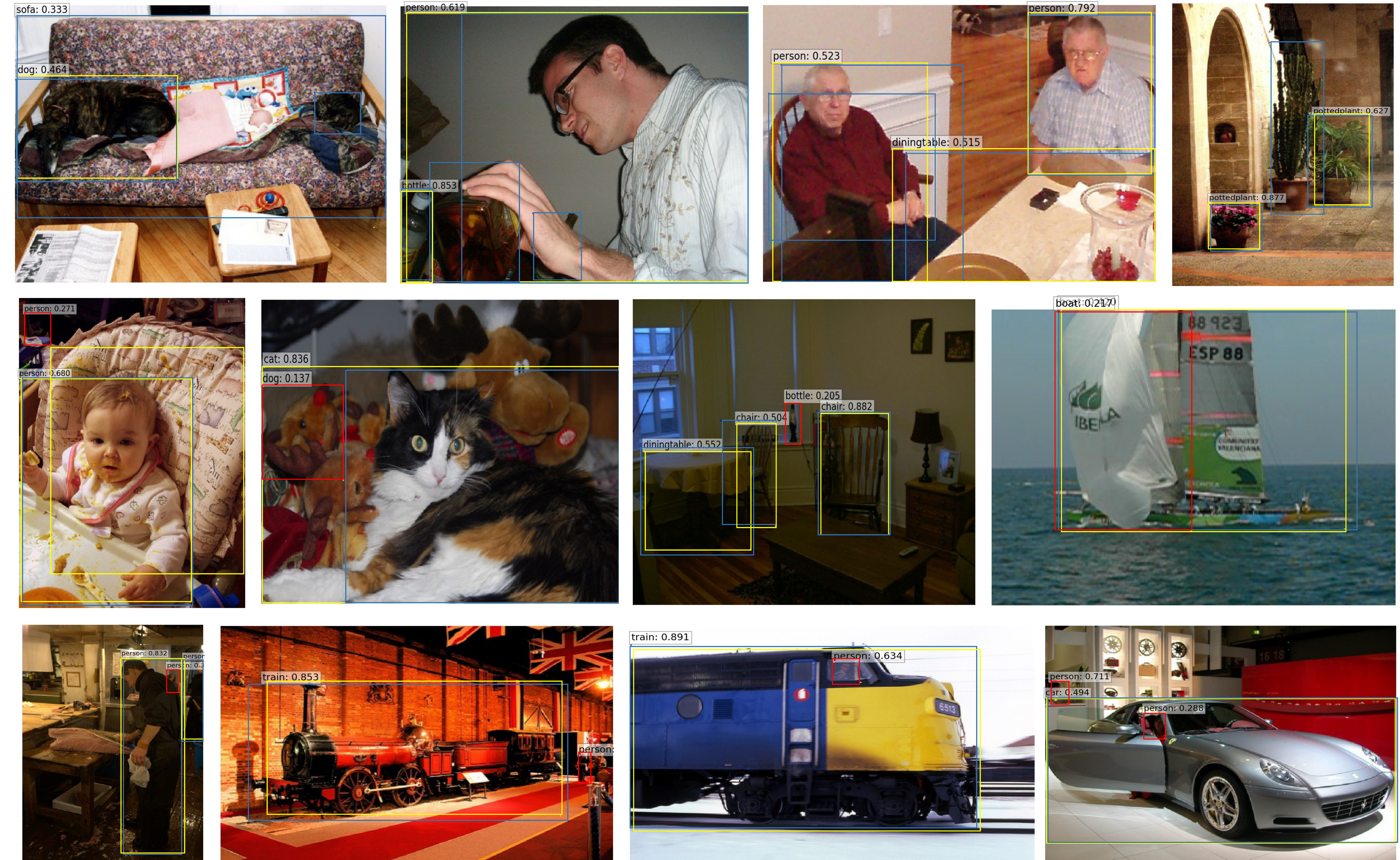}\\
  \caption{Visualization of the detection results on PASCAL VOC 2007. The ground-truth of objects are drawn in blue color, the true positives are drawn in yellow color, and the false positives are drawn in red color. Zoom in to find the ``false" false positives. }    \label{fig:vocError}
\end{figure*}

Some detection results on this dataset are shown in Fig.~\ref{fig:vocError}. As it shown, PLN640 can accurately detect object in a wide range of scales. In this figure, the first row shows some hard examples missed by our detector, the second row shows false positives, and the third row shows some interesting ``false" false positives found by PLN640, which are actually true positives, however they are not labeled in ground-truth, \eg, the small or heavily occluded person and bottle.

\subsubsection{Ablation studies}

In this section, we carry out ablation studies, which show the progress of how we designed the PLN detector. All the experiments are carried out on the PASCAL VOC 2007 dataset. The experiment results are given in Table~\ref{table:analysis}.

\begin{table}[ht!]
\caption{Effects of various design choices and components on the performance of PLN detector. The results are tested on the PASCAL VOC 2007 dataset. Except the row of ``higher resolution'', the results are given by PLN448.}
  \centering
  \resizebox{0.5\textwidth}{!}{
  
  \begin{tabular}{c|c c c c c c  c}
  \toprule
    & \multicolumn{6}{|c}{PLN} \\
  \midrule
    left-top/right-bottom pair     & $\surd$ & $\surd$ &         &         &         &         &         \\
    center/left-top pair           &         &         & $\surd$ & $\surd$ & $\surd$ & $\surd$ & $\surd$ \\
    multi points per-grid     &         & $\surd$ & $\surd$ & $\surd$ & $\surd$ & $\surd$ & $\surd$ \\
    larger receptive field         &         &         &         & $\surd$ & $\surd$ & $\surd$ & $\surd$ \\
    data-aug for small size obj    &         &         &         &         & $\surd$ & $\surd$ & $\surd$ \\
    merging four branches          &         &         &         &         &         & $\surd$ & $\surd$ \\
    higher resolution (640$\times$640)             &         &         &         &         &         &         & $\surd$ \\
   \midrule
    mAP                            &  71.9   &  72.1   &  73.2   &  74.4   &  75.8   &  76.8   &  78.8   \\
  \bottomrule
  \end{tabular}
  
  }
  \label{table:analysis}
\end{table}

\textbf{Predicting center-corner pair is better than corner-corner pair.} Compared with the previous detectors, we find object of a given image by predicting point pair of object. In the beginning, we use deep ConvNets to predict left-top and right-bottom point. Then, we realize that bounding box can also be determined by a center-corner pair and center point is much easier to be localized than corner point. Therefore we change to predict center and left-top point, which makes our method improve 1.1\% mAP as we expected. Please refer to the second column and the third column in Table~\ref{table:analysis}. The reason why center point is easier to be localized by ConvNet is because it is inside the object and has richer information around.

\textbf{Predicting multiple point pairs per-grid helps.} Inspired by the design of predicting multiple bounding boxes per-grid in YOLO \cite{Ref:YOLO-Redmon2016}, we predict multiple point pairs to increase the recall of PLN detector. Specifically, we set $B=2$ in Section~\ref{sec:design}. It just improves 0.2\% mAP using when using the left-top/right-bottom point pair. Please refer to the first column and the second column.

\textbf{Larger receptive field of the convolutional feature map is beneficial.} Though the points could be localized using local information, it needs large receptive fields to predict the object category and the linked point. In order to obtain larger receptive field compared to the original Inception-v2, we apply dilation convolution to the convolutional feature map as the last convolutional layer in PLN, as shown in Fig~\ref{fig:network}. It does not increase much computation because of the low dimension of feature map which specified by the detection parameters (such as, number of object classes and number of grid cells). \emph{Different from classification, detection needs more detail, so we also fuse a high resolution Inception block feature map to the last inception block}. After that, we obtain a 1.2\% mAP improvement. Please refer to the third column and the fourth column.

\textbf{Data augmentation of small size objects} In the task of classification, we crop large image patches to augment dataset, it always relatively enlarges objects in the images. When using these data to train our network, it helps to detect large objects, but it decreases the performance of detecting small objects. We use a randomly switched data augmentation of YOLO and SSD, but the proposed data augmentation method tends to jitter outside of the cropped image patches to make the objects inside smaller. It improves performance of the system by 1.4\% mAP.

\textbf{Merging multiple branches is beneficial.} Occlusion is a big challenge in object detection. PLN detector using a single point pair may be easily affected by corner point missing. However, the other corner points may be visiable. Therefore, we use four branches to predict four different center-corner points pairs. In the testing, we merge the bounding boxes of the four branches, then apply NMS to obtain the final object detection results. It improves 1.0\% mAP. We also test the mAP of four branches separately, and the results are given in Table~\ref{table:branch}. The results show that the four corner points give similar performance and merging them gives obvious performance gain.

\begin{table}[ht!]
\caption{Detection performance of different branches and the merged results.}
  \centering
  \begin{tabular}{l|c}
  \toprule
    & \multicolumn{1}{|c}{mAP} \\
    \midrule
    left-top branch    &  75.4    \\
    right-top branch   &  75.7    \\
    left-bottom branch  &  75.5    \\
    right-bottom branch &  75.2    \\
    \hline
    Merging branches &  76.8    \\
    \bottomrule
  \end{tabular}
  
  \label{table:branch}
\end{table}


\textbf{Higher resolution of input image is beneficial to detection.} Object detection requires larger input image for preserving the details to localize small objects. Fast and Faster R-CNN take input images whose minimum dimension is 600. SSD shows that using image resolution of 512 is better than 300. In our experiments, we compare three different input sizes, including 448, 512 and 640. The results can be observed from Table~\ref{table:analysis}. It shows that PLN detector obtains 1.0\% mAP improvement by increasing the input image size both from 448$\times$448 to 512$\times$512 and  from 512$\times$512 to 640$\times$640.

\subsubsection{Error Analysis}


We use the methodology and tools of Hoiem \etal \cite{hoiem2012diagnosing} to analyze the detection results of PLN448. The error types include localization error (correct class, .1$<$IoU$<$.5), similar class error (mis-classified to the similar classes, IoU$>$.1), other class error (mis-classified to the other classes, IoU$>$.1) and background error (IoU$<$.1). For more details about this tool, please refer to \cite{hoiem2012diagnosing}.

\begin{figure}[ht!]
  \centering
  \includegraphics[width=1.0\linewidth]{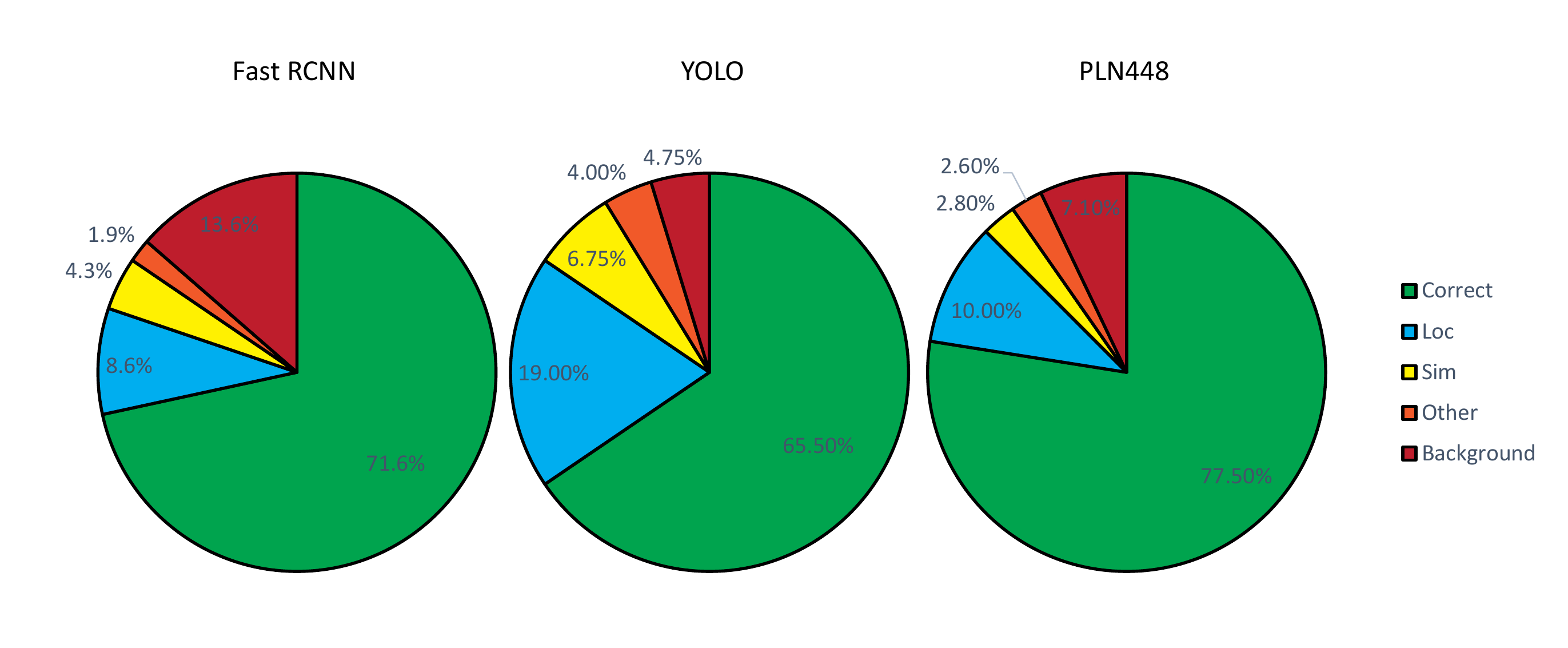}\\
  \caption{{Error analysis On PLN, Faster R-CNN~\cite{Ref:FasterRCNN-Ren2015} and YOLO~\cite{Ref:YOLO-Redmon2016}.} }    \label{fig:pie}
\end{figure}

From the pie chart in Fig.~\ref{fig:pie}, PLN has fewer location error than YOLO, and just a little worse than Faster R-CNN. It does not be influenced by background clutters seriously as YOLO. Overall, PLN is more balanced on different types of errors. We can also find that location error and background error are the main factors that affect the performance of the three detectors.

\subsubsection{Speed}

In the object detection papers, different detectors tend to use different base models. YOLOv2 uses Darknet19, SSD adopts VGG16, while Faster R-CNN employs ResNet101. According to the survey paper by Google researchers \cite{huang2016speed}, Inception-V2 is a good trade-off speed/accuracy and has fewer FLOPs than VGG16 and ResNet101. 

However, except the base net, we have additional conv layers in the four branches and the inference computations as shown in Fig.~\ref{fig:network}, which makes PLN slower than YOLOv2 and SSD. The original PLN448 has a speed of 10fps on a GTX 1080 GPU using an unoptimized python implementation, in which the time taken by Inception-V2, additional conv layers and inference are 13\%, 84\% and 3\%, respectively. Then, we remove a 3*3 conv layer and replace four 3*3 conv layers with 1*1 conv layers, in which the time taken by Inception-V2, additional conv layers and inference are 32\%, 62\% and 6\%, respectively. This simplified PLN448 obtains 24fps with 0.5 mAP accuracy drops on VOC 2007. The results show that the most time costing part is the additional conv layers. We will further study how to reduce the computation using small filters and residual connections to pursue the speed of SSD/YOLO. 

\subsection{PASCAL VOC 2012}

In the experiments on PASCAL VOC 2012, we follow the standard protocol to use VOC 2012 trainval, VOC 2007 trainval and VOC 2007 test for training, and use VOC 2012 test for testing. We use the same hyper-parameters as used in the VOC 2007 experiments. Our detection results are uploaded to the evaluation server and the mAPs are reported in Table~\ref{table:voc12} and compared with the other methods.

In Table~\ref{table:voc12}, we show the detection results of 
PLN512\footnote{\url{http://host.robots.ox.ac.uk:8080/anonymous/TDF0RU.html}} and PLN640\footnote{\url{http://host.robots.ox.ac.uk:8080/anonymous/VWP68I.html}} on PASCAL VOC 2012. The results show that PLN has obtained significantly better average precision on most of the difficult classes, such as boat, bottle and chair. The mAP over 20 classes is 0.760, which is the state-of-the-art. In addition, the results also show that increasing image size from 512 to 640 has consistent performance gain as observed in PASCAL VOC 2007 dataset.

\begin{table*}
\caption{{PASCAL VOC 2012 test dectection results.} 
  The image size for YOLOv2 is 544 $\times$ 544, 
  Faster R-CNN use images with minimum dimension 600, 
  the two SSD models have different input sizes (300 $\times$ 300 vs. 512 $\times$ 512)
  and our PLN models have input sizes 512 $\times$ 512 and 640 $\times$ 640. }
  \centering
  \resizebox{\textwidth}{!}{\begin{tabular}{l|c|c c c c c c c c c c c c c c c c c c c c}
  \toprule
    Method & mAP & aero & bike & bird & boat & bottle & bus & car & cat & chair & cow & table & dog & horse & mbike & person & plant & sheep & sofa & train & tv \\
    \midrule
    Faster        & 73.8 & 86.5 & 81.6 & 77.2 & 58.0 & 51.0 & 78.6 & 76.6 & 93.2 & 48.6 & 80.4 & 59.0 & 92.1 & 85.3 & 84.8 & 80.7 & 48.1 & 77.3 & 66.5 & 84.7 & 65.6 \\
    SSD300        & 72.4 & 85.6 & 80.1 & 70.5 & 57.6 & 46.2 & 79.4 & 76.1 & 89.2 & 53.0 & 77.0 & 60.8 & 87.0 & 83.1 & 82.3 & 79.4 & 45.9 & 75.9 & 69.5 & 81.9 & 67.5 \\
    SSD512        & 74.9 & 87.4 & 82.3 & 75.8 & 59.0 & 52.6 & 81.7 & 81.5 & 90.0 & 55.4 & 79.0 & 59.8 & 88.4 & 84.3 & 84.7 & 83.3 & 50.2 & 78.0 & 66.3 & 86.3 & 72.0 \\
    YOLOv2        & 73.4 & 86.3 & 82.0 & 74.8 & 59.2 & 51.8 & 79.8 & 76.5 & 90.6 & 52.1 & 78.2 & 58.5 & 89.3 & 82.5 & 83.4 & 81.3 & 49.1 & 77.2 & 62.4 & 83.8 & 68.7 \\
    \midrule
    PLN512        & 75.2 & 87.1 & 84.5 & 75.7 & 63.6 & 52.2 & 81.4 & 76.5 & 91.9 & 56.0 & 78.6 & 60.3 & 88.8 & 85.6 & 84.4 & 81.3 & 54.6 & 78.5 & 66.6 & 86.7 & 70.2 \\
    PLN640        & 76.0 & 88.3 & 84.7 & 77.4 & 66.0 & 55.8 & 82.1 & 79.4 & 91.9 & 58.2 & 77.3 & 58.8 & 89.5 & 85.3 & 85.3 & 82.9 & 55.8 & 79.6 & 64.6 & 86.5 & 69.9 \\
    \bottomrule
  \end{tabular}}
  \label{table:voc12}
\end{table*}

\begin{figure*}
  \centering
    \includegraphics[width=0.8\linewidth]{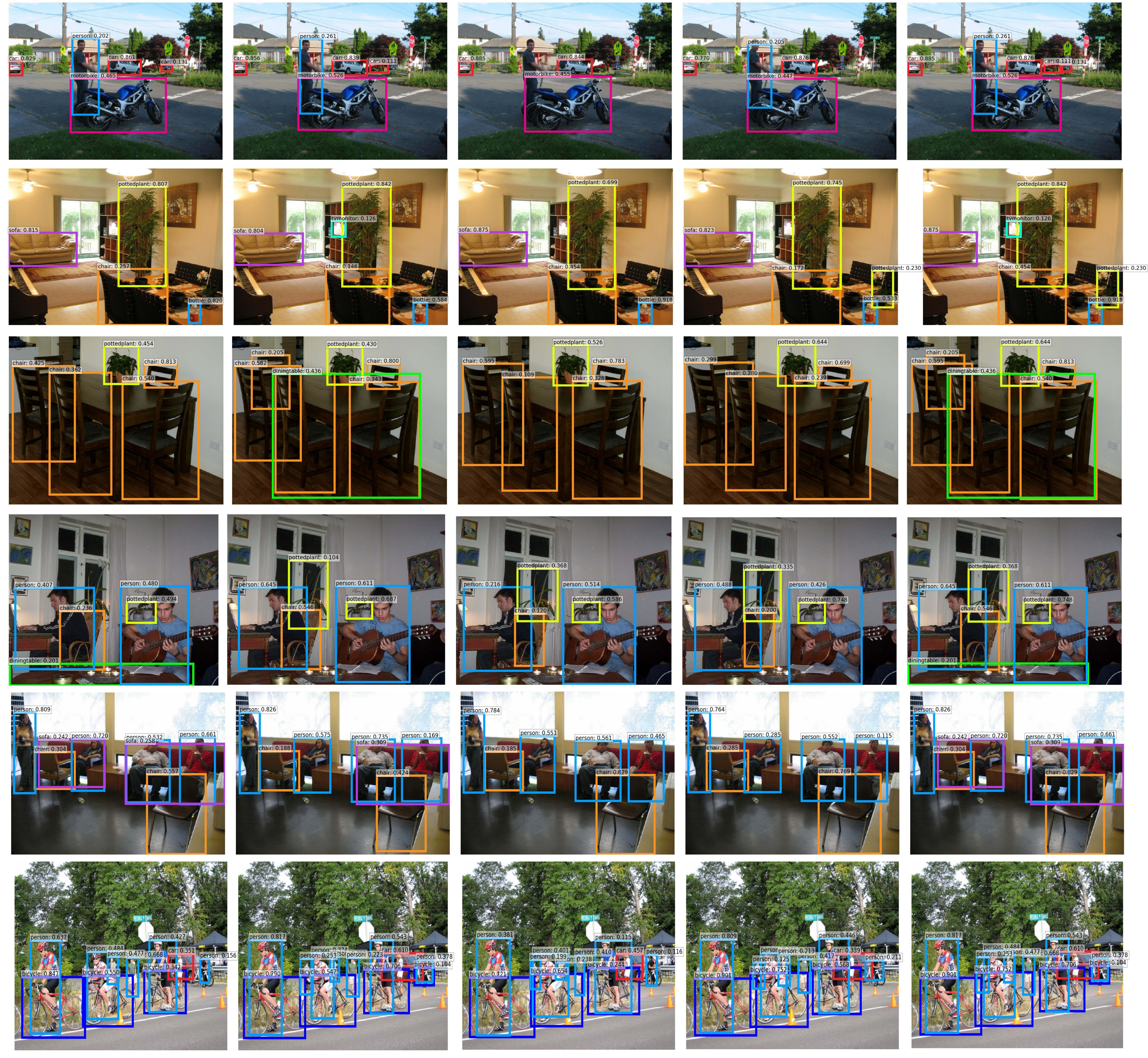}\\
  \caption{\textbf{Comparison the detection results using different corner points paired with the center point.} From left to right, each column shows the detection results of using the left-top corner, the right-top corner, the left-bottom corner, the right-bottom corner and the results by fusing the four corners. }    \label{fig:corners}
\end{figure*}

We also show a comparison of detection results using different corner points and their merged results in Fig.~\ref{fig:corners}. As shown in the figure, the plotted plant in the 2nd row can only be detected using the right-bottom corner and the dining table can only be detected using the left-top corner; however, they are all correctly detected in the merged results. The results show that the fused results are more robust to occlusion as claimed in the introduction section.

\subsection{People-Art Dataset}

We also verify our algorithm on People-Art dataset, which only consists of people, in 43 different styles. People in this dataset is quite different from those in common photographs. We train our model only on PASCAL VOC 2010 using 448 input size, and test on the test dataset of People-Art. We first train our model with 0.001 learning rate for 200 iterations, and then increase learning rate to 0.005 for 3w iterations. Throughout training, we use a batch size of 40, a momentum of 0.9, and a decay of 0.00004. Other settings are same as PASCAL VOC. Thus we train our model with 20 categories and use this model to detect people of People-Art dataset. From the Table~\ref{table:people}, it can be observed that our method can generalize well on dataset of other domain.

\begin{table}[ht!]
\caption{ Detection results on People Art dataset. }
  \centering
  \begin{tabular}{c|cccc}
  \toprule
    &DPM~\cite{felzenszwalb2010object}&R-CNN~\cite{Ref:RCNN-Girshick2014}&YOLO~\cite{Ref:YOLO-Redmon2016}&PLN448\\
    \midrule
    AP&  32 &  26 &  45 &  47    \\
    \bottomrule
  \end{tabular}
  \label{table:people}
\end{table}

\subsection{COCO dataset}

To further validate PLN, we evaluate it on the COCO dataset, which is the most challenging object detection benchmark to date. We use COCO trainval to train PLN512. The trainval set contains about 120k training images.
We first train our model with 0.001 learning rate for 400 iterations, and then increase learning rate to 0.005 for 700k iterations. Throughout training, we use a batch size of 50, a momentum of 0.9, and a decay of 0.00004. Other settings are same as the experiment setting of PASCAL VOC.

Table~\ref{table:coco} shows the detection results on COCO test-dev2015 of PLN512 compared with YOLOv2 \cite{Ref:YOLO9000-Redmon2017}, SSD \cite{Ref:SSD-Redmon2016}, ION \cite{Bell2016Inside} and Faster R-CNN \cite{Ref:FasterRCNN-Ren2015}. PLN512 achieves best detection result in both mAP@[0.5:0.95] and mAP@0.5 among the compared methods in the table. 

We focus on comparing PLN512 with the single-shot detectors, YOLOv2 and SSD512, since the three detectors have similar base network structure. The results show than PLN512 is significantly better than YOLOv2, which proves the effectiveness of the proposed detection loss function. Even without the multi-scale predictions, we can find that PLN512 is still better than SSD512 in both mAP@0.5 and mAP@[0.5:0.95], which is not a trivial achievement.


\begin{table*}[ht!]
\caption{COCO test-dev2015 detection results}
  \centering
  \begin{tabular}{c|c|c c c|c c c|c c c|c c c}
  \toprule
    & & \multicolumn{3}{|c}{Avg.Precision, IoU:} & \multicolumn{3}{|c}{Avg.Precision, Area:} & \multicolumn{3}{|c}{Avg.Recall, \#Dets:} & \multicolumn{3}{|c}{Avg.Recall, Area} \\
    Method & data & 0.5:0.95 & 0.5 & 0.75 & S & M & L & 1 & 10 & 100 & S & M & L \\
    \midrule
    Fast \cite{Ref:FastRCNN-Girshick2015}   & train       & 19.7 & 35.9 & -    & -   & -    & -    & -    & -    & -    & -    & -    & -    \\
    Fast \cite{Bell2016Inside}  & train       & 20.5 & 39.9 & 19.4 & 4.1 & 20.0 & 35.8 & 21.3 & 29.5 & 30.1 & 7.3  & 32.1 & 52.0 \\
    Faster \cite{Ref:FasterRCNN-Ren2015} & trainval    & 21.9 & 42.7 & -    & -   & -    & -    & -    & -    & -    & -    & -    & -    \\
    ION \cite{Bell2016Inside}   & train       & 23.6 & 43.2 & 23.6 & 6.4 & 24.1 & 38.3 & 23.2 & 32.7 & 33.5 & 10.1 & 37.7 & 53.6 \\
    Faster \cite{lin2014microsoft} & trainval    & 24.2 & 45.3 & 23.5 & 7.7 & 26.4 & 37.1 & 23.8 & 34.0 & 34.6 & 12.0 & 38.5 & 54.4 \\
    SSD300 \cite{Ref:SSD-Redmon2016} & trainval35k & 23.2 & 41.2 & 23.4 & 5.3 & 23.2 & 39.6 & 22.5 & 33.2 & 35.3 & 9.6  & 37.6 & 56.5 \\
    SSD512 \cite{Ref:SSD-Redmon2016} & trainval35k & 26.8 & 46.5 & 27.8 & 9.0 & 28.9 & 41.9 & 24.8 & 37.5 & 39.8 & 14.0 & 43.5 & 59.0 \\
    YOLOv2 \cite{Ref:YOLO9000-Redmon2017} & trainval35k & 21.6 & 44.0 & 19.2 & 5.0 & 22.4 & 35.5 & 20.7 & 31.6 & 33.3 & 9.8  & 36.5 & 54.4 \\
    \midrule
    PLN512 & trainval    & 28.9 & 48.3 & 29.4 & 8.1 & 30.6 & 46.8 & 26.9 & 41.7 & 45.0 & 15.1 & 49.7 & 70.9 \\
    \bottomrule
  \end{tabular}
  \label{table:coco}
\end{table*}

\section{Conclusion}

In this paper, a novel deep object detection network is proposed. It is totally a new object detection framework and different from the sliding windows, object proposal, and bounding boxes regression frameworks. In this new framework, object detection is done by linking the center point and corner point of object. Thus, it is robust to object occlusion, scale variation, and aspect ratio variation. The point linking and point detection are implemented in a single deep network trained in an end-to-end manner. The superior results compared to Faster R-CNN with deep residual networks, the latest YOLOv2 and SSD512 on PASCAL VOC and COCO have proven the effectiveness of the proposed PLN method.  

\section*{Acknowledgement}
This work is supported by Young Elite Scientists Sponsorship (YESS) Program by CAST (No. YESS20150077) and NSFC 61503145. We thank Zhuowen Tu for helpful discussions.

{\small
\bibliographystyle{ieee}
\bibliography{PL}
}

\end{document}